\documentclass[conference]{IEEEtran}
\IEEEoverridecommandlockouts

\usepackage{amsmath,graphicx}
\usepackage{cite}
\usepackage{amsfonts}
\usepackage{textcomp}
\usepackage{xcolor}
\usepackage{booktabs}
\usepackage{color}
\usepackage{xcolor}
\usepackage{subfigure}
\usepackage{amsmath}
\usepackage{algorithm}
\usepackage{algpseudocode}
\usepackage{amssymb}  
\usepackage{multirow}
\usepackage{hyperref}
\usepackage{subcaption}

\def\BibTeX{{\rm B\kern-.05em{\sc i\kern-.025em b}\kern-.08em
    T\kern-.1667em\lower.7ex\hbox{E}\kern-.125emX}}
\begin{document}

\title{
Federated Domain Generalization with Label Smoothing and Balanced Decentralized Training
}

\author{
\IEEEauthorblockN{Milad Soltany\IEEEauthorrefmark{1}, Farhad Pourpanah\IEEEauthorrefmark{1}, Mahdiyar Molahasani\IEEEauthorrefmark{1}, Michael Greenspan, and Ali Etemad
\IEEEauthorblockA{Department of Electrical and Computer Engineering\\ Queen's University, Kingston, Canada
\\\{milad.soltany, f.pourpanahnavan, m.molahasani, michael.greenspan, ali.etemad\}@queensu.ca
}
}
}

\maketitle

\let\thefootnote\relax  
\footnotetext{*These authors contributed equally.}
\begin{abstract}
In this paper, we propose a novel approach, Federated Domain Generalization with Label Smoothing and Balanced Decentralized Training (FedSB), to address the challenges of data heterogeneity within a federated learning framework. FedSB utilizes label smoothing at the client level to prevent overfitting to domain-specific features, thereby enhancing 
generalization capabilities across diverse domains
when aggregating local models into a global model. Additionally, FedSB incorporates a decentralized budgeting mechanism
which
balances training among clients, 
which is shown to improve the performance of the aggregated global model. 
Extensive experiments on four commonly used multi-domain datasets, PACS, VLCS, OfficeHome, and TerraInc,
demonstrate that FedSB outperforms competing methods, achieving state-of-the-art results on three out of four datasets,
indicating the effectiveness of FedSB in addressing
data heterogeneity.

\end{abstract}

\section{Introduction}
\label{sec:intro}

Federated learning (FL) \cite{mcmahan2017communication}
has recently gained interest in machine learning (ML) as an alternative to centralized training in scenarios where there are either data privacy concerns or resource-sharing limitations. Unlike centralized ML methods, which requires data from multiple sources to be accessed in one global model \cite{Zhang_2022_CVPR,zhou2020learning,wang2020learning, xu2021fourier}
, FL allows for collaborative training of local models without sharing data or resources among them, thus preserving both data privacy and security. The parameters of the locally-trained models are occasionally uploaded to a central server, where they are aggregated to form a unified global model. For instance, a scenario has been presented in \cite{nguyen2022fedsr}, where multiple hospitals collaborate to train a diagnostic model while ensuring that their patients' data remains private to each hospital and is not shared.

An underlying challenge in FL is data heterogeneity, where the data available to each local client follows different distributions, making it difficult for the global model to generalize effectively to \textit{unseen} target domains.
To address this challenge, federated domain generalization (FDG) methods \cite{bai2023benchmarking,zhang2023federated} have been proposed. These methods mostly focus on either learning domain-invariant features \cite{nguyen2022fedsr} or learning common features across domains \cite{zhang2021federated} to enhance the global model's robustness across diverse environments. 
While FDG techniques show promise in tackling data heterogeneity, two challenges persist as a result of this phenomenon.
The first is the overconfidence of the clients on their local data, as they tend to learn domain-specific features. This overconfidence limits the effectiveness of these local models when aggregated to form the global model. 

Secondly, the distribution of samples varies across clients, causing imbalances in model training, where clients with more samples contribute more to the global model.
This in turn can result in a biased or sub-optimal performance \cite{li2020federated}.

In this paper, we propose 
\textit{\textbf{Fed}erated Domain Generalization with Label \textbf{S}moothing and \textbf{B}alanced Decentralized Training} (FedSB) to address these issues. 
FedSB mitigates local model overconfidence by using label smoothing to promote domain-invariant representation learning, enhancing the model's generalization across different domains. Additionally, we introduce a simple and innovative budgeting mechanism that allocates training resources according to each client's local data volume, 
thereby ensuring balanced and consistent training contributions from all clients which ultimately improves the overall model performance. Our approach is designed to enhance the robustness and generalization of FL models across diverse and unseen domains. We conduct comprehensive experiments, ablation studies, and sensitivity analyses to evaluate the effectiveness of FedSB using four datasets from the domainbed benchmark\cite{gulrajani2021in}: PACS 
\cite{li2017deeper}
, OfficeHome 
\cite{venkateswara2017deep}
, TerraInc 
\cite{beery2018recognition}
, and VLCS 
\cite{fang2013unbiased}
, achieving state-of-the-art performance on 3 of the 4 datasets.

\begin{figure*}
    \centering
    \includegraphics[width=\linewidth]{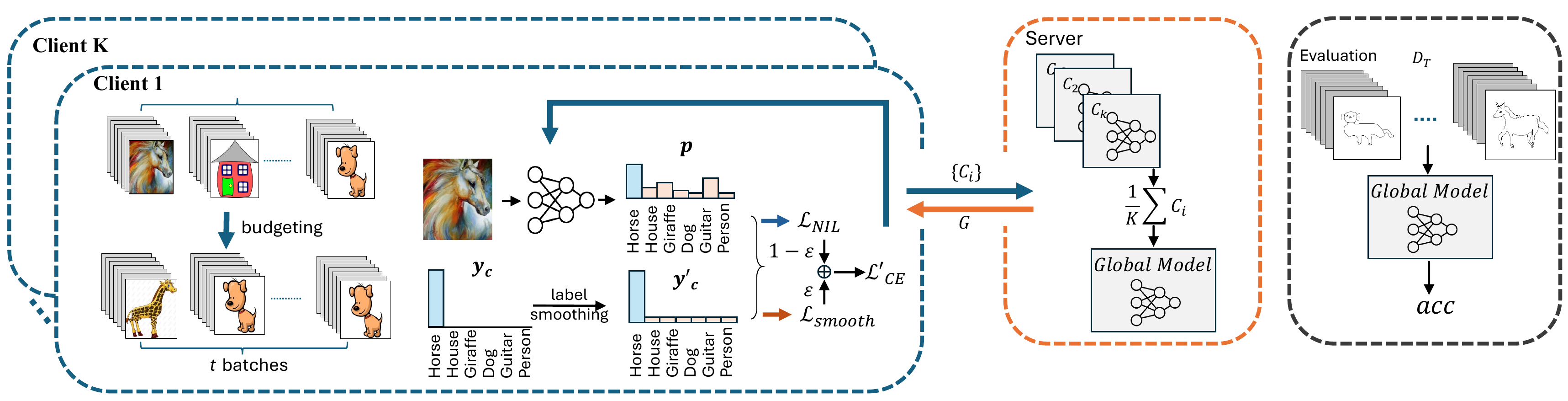}
    \caption{\textbf{Overview of FedSB.} 
    \textcolor{blue}{Blue} and \textcolor{orange}{Orange} arrows show local model loading to the server and initialization with the global model from the server, respectively. }
    \label{fig:overview}
\end{figure*}

Our contributions in this paper are: (\textbf{1}) We propose label smoothing as an effective means to address the issue of overconfidence in local clients, leading to improved generalization across unseen domains. To the best of our knowledge, this is the first application of label smoothing in the context of federated domain generalization. 
(\textbf{2}) We propose a novel budgeting technique 
to effectively mitigate the data heterogeneity challenge faced by local clients. (\textbf{3}) We evaluate FedSB on multiple domain generalization datasets, achieving state-of-the-art performances. We also conduct comprehensive ablation studies and sensitivity analyses to demonstrate the effectiveness of our approach. To enable fast and accurate reproducibility, we make the code public at: \href{https://github.com/miladsoltany/FedSB}{https://github.com/miladsoltany/FedSB}.

\section{Related Work}
\label{sec:related}

\noindent \textbf{Federated learning.}
FL aims to collaboratively optimize a central model through training multiple local clients, e.g., devices or organizations, without sharing data to preserve privacy \cite{mcmahan2017communication, konevcny2016federated}. FedAVG \cite{mcmahan2017communication} is a very popular algorithm, where local models are iteratively uploaded to a central server, and then through a weighted averaging scheme, a global model is produced, which will then be downloaded to the clients. Many other methods build on top of FedAVG to enhance its capabilities and performance. For instance, FedProx \cite{li2020federated} adds a proximal term to the local loss to prevent the local models from deviating extensively from the global model. MOON \cite{li2021model} utilizes model contrastive learning to correct local representations. Federated Dropout \cite{wen2022federated} introduces dropout techniques into FL to reduce the impact of system heterogeneity. FedNova \cite{wang2020tackling} proposes a normalized averaging technique to address system heterogeneity by accounting for differences in the number of local updates each client performs.

\noindent \textbf{Federated domain generalization.}
In FDG, the primary goal is to train a global model through multiple local clients collaboratively. This model should perform well in an unseen domain.  A key feature of FDG is its ability to achieve this without explicitly sharing data among the clients, thereby preserving data privacy. Most works in this area utilize an FL setup, where each client has access to only one data domain, and a central server is responsible for aggregating these local clients to produce one unified model. FedSR \cite{nguyen2022fedsr} employs L2 norm and conditional mutual information regularizers at the client level to discourage the learning of domain-specific features. 
Some recent works share some form of information amongs clients, which could potentially lead to breach of data privacy. CCST \cite{chen2023federated} facilitates the sharing of a style bank between clients, where the style bank comprises the mean and standard deviation of representations generated from each client's local data using a CNN-based backbone, such as VGG 
. A style transfer model
is then employed to transfer the styles of images from one client to match the styles of another, promoting cross-client generalization. COPA \cite{wu2021collaborative} shares the classification heads among the clients. Through this, the local encoders are encouraged to learn domain invariant features from their local data that work well even with classification heads from other domains. This comes at the cost of communicating extra layers with each communication round. FedProto \cite{tan2022fedproto} and FedCDG \cite{yu2023contrastive} take a different approach by sharing class prototypes among clients. This effectively distributes class representations across clients, but it overlooks the fact that explicit information sharing can undermine privacy.
Additionally, FedGaLa \cite{pourpanah2024federated} goes a step further and extends FDG to the unsupervised realm, where the objective is to learn domain-invariant representations from unlabeled samples through local and global gradient alignment.

\section{Method}
\noindent\textbf{Problem Formulation.}
Let's assume $i \in \{1,2,\cdots,K\}$, where $K$ is the total number of clients in the system. Here, client $\mathcal{C}_i$ has access to its local dataset $\mathcal{D}_i = \{x_i, y_i\}$ from a specific distribution $p_i(x,y)$, with $x\!\in\!X$ and $y\!\in\!Y$ representing the inputs and labels, respectively. The objective of FDG is to learn a global model $\Theta$ by aggregating the local models $\theta_i$, such that the global model can generalize to an unseen target domain with distribution $p_T(x,y)$, where $p_T(x)\!\neq\!p_i(x)$ for $i \in [1,K]$, and ${\cal{D}}_{T}$ is not available during training.



\begin{table*}[ht]
\caption{Comparison of image recognition accuracy on PACS, OfficeHome, TerraIncognita and VLCS datasets. The single-letter columns represent the unseen (test) domain in the corresponding dataset.}
\setlength{\tabcolsep}{3pt}
\label{tab:results}
\begin{center}
\resizebox{\textwidth}{!}{%
\begin{tabular}{lccccclccccclccccclccccc}
\toprule
 & \multicolumn{5}{c}{PACS} &  & \multicolumn{5}{c}{OfficeHome} &  & \multicolumn{5}{c}{TerraIncognita} &  & \multicolumn{5}{c}{VLCS} \\ \cmidrule{2-24}
 Method & P & A & C & S & Ave. &  & P & A & C & R & Ave. &  & L36 & L43 & L48 & L100 & Ave. &  & V & L & C & S & Ave. \\ \midrule
 
FedAvg \cite{mcmahan2017communication} & 91.67\textbf{$\pm$}\scriptsize{0.9} & 79.25\textbf{$\pm$}\scriptsize{1.3} & 70.46\textbf{$\pm$}\scriptsize{0.8} & 75.98\textbf{$\pm$}\scriptsize{1.7} & 79.34 &
& 78.30\textbf{$\pm$}\scriptsize{1.0} & 64.16\textbf{$\pm$}\scriptsize{0.7} & 54.88\textbf{$\pm$}\scriptsize{0.6} & 79.08\textbf{$\pm$}\scriptsize{0.4} & \underline{69.10}  &
&41.37\textbf{$\pm$}\scriptsize{5.2} &43.34\textbf{$\pm$}\scriptsize{3.7} &39.90\textbf{$\pm$}\scriptsize{4.8} &56.02\textbf{$\pm$}\scriptsize{3.2} &45.16 &  
&72.27\textbf{$\pm$}\scriptsize{2.9}  &60.25\textbf{$\pm$}\scriptsize{3.5}  &96.35\textbf{$\pm$}\scriptsize{1.2}  &70.37\textbf{$\pm$}\scriptsize{2.1}  &74.81 \\
 
FedADG \cite{zhang2021federated}  &93.64 &81.39 &75.39 &78.56 &82.25 &  
&74.87 &60.27 &\underline{56.09} &76.48 &66.93 &  
&-  & - & - &- &- &  
&73.20  &61.20  &95.78  &74.95  &76.28  \\ 
 
FedProx \cite{li2020federated} &91.69\textbf{$\pm$}\scriptsize{1.3} &79.16\textbf{$\pm$}\scriptsize{0.6} &71.45\textbf{$\pm$}\scriptsize{3.0} &75.51\textbf{$\pm$}\scriptsize{3.5} &79.45 &
&\underline{77.45}\textbf{$\pm$}\scriptsize{0.3} &\underline{64.33}\textbf{$\pm$}\scriptsize{0.6} &55.02\textbf{$\pm$}\scriptsize{0.5} &\textbf{79.53}\textbf{$\pm$}\scriptsize{0.1} &68.14 &  
&\textbf{43.13}\textbf{$\pm$}\scriptsize{2.1}  &\textbf{45.79}\textbf{$\pm$}\scriptsize{3.2}  &40.64\textbf{$\pm$}\scriptsize{4.6}  &55.18\textbf{$\pm$}\scriptsize{2.7}  &46.19  &
&73.16\textbf{$\pm$}\scriptsize{2.0}  &60.39\textbf{$\pm$}\scriptsize{2.4}  &97.03\textbf{$\pm$}\scriptsize{0.9}  &72.62\textbf{$\pm$}\scriptsize{2.8}  &75.80  \\
 
 
FedSR \cite{nguyen2022fedsr}  &93.00\textbf{$\pm$}\scriptsize{1.4} &78.37\textbf{$\pm$}\scriptsize{1.6} &72.22\textbf{$\pm$}\scriptsize{2.4} &77.53\textbf{$\pm$}\scriptsize{2.9} &80.27 &
&73.50\textbf{$\pm$}\scriptsize{0.7} &58.46\textbf{$\pm$}\scriptsize{0.1} &50.83\textbf{$\pm$}\scriptsize{0.1} &73.93\textbf{$\pm$}\scriptsize{1.1} &64.18  & 
&24.33\textbf{$\pm$}\scriptsize{8.8}  &33.43\textbf{$\pm$}\scriptsize{5.9}  &30.97\textbf{$\pm$}\scriptsize{5.2}  &56.80\textbf{$\pm$}\scriptsize{3.1} &36.38 &
&68.83\textbf{$\pm$}\scriptsize{3.6}  &59.63\textbf{$\pm$}\scriptsize{4.3}  &95.83\textbf{$\pm$}\scriptsize{3.2}  &71.63\textbf{$\pm$}\scriptsize{1.8}  &73.98  \\
 
FedIIR \cite{guo2023out}  &\textbf{94.56}\textbf{$\pm$}\scriptsize{0.9} &80.06\textbf{$\pm$}\scriptsize{0.8} &75.20\textbf{$\pm$}\scriptsize{1.7} &\underline{79.63}\textbf{$\pm$}\scriptsize{1.65} &\underline{82.36} &
&75.68\textbf{$\pm$}\scriptsize{0.5} &63.57\textbf{$\pm$}\scriptsize{0.4} &54.53\textbf{$\pm$}\scriptsize{0.5} &78.16\textbf{$\pm$}\scriptsize{0.2} &68.08  &  
&41.10\textbf{$\pm$}\scriptsize{6.2}  &47.79\textbf{$\pm$}\scriptsize{4.5}  &39.47\textbf{$\pm$}\scriptsize{3.6}  &47.63\textbf{$\pm$}\scriptsize{2.4} &44.01  &  
&\textbf{76.42}\textbf{$\pm$}\scriptsize{1.6} &61.84\textbf{$\pm$}\scriptsize{3.1}   &96.79\textbf{$\pm$}\scriptsize{0.6}  &74.69\textbf{$\pm$}\scriptsize{3.7}  &\textbf{77.44}  \\
 
 
FedSB (ours)&\underline{94.19}\textbf{$\pm$}\scriptsize{0.1} &\textbf{81.80}\textbf{$\pm$}\scriptsize{0.7} &\underline{75.28}\textbf{$\pm$}\scriptsize{2.9} &\textbf{83.52}\textbf{$\pm$}\scriptsize{0.9} &\textbf{83.81} &
&\textbf{78.33}\textbf{$\pm$}\scriptsize{0.3} &\textbf{65.88}\textbf{$\pm$}\scriptsize{0.3} &\textbf{60.05}\textbf{$\pm$}\scriptsize{0.7} &\underline{79.42}\textbf{$\pm$}\scriptsize{0.4} &\textbf{70.92}  &  
&38.37\textbf{$\pm$}\scriptsize{4.8}  &44.56\textbf{$\pm$}\scriptsize{4.1}  &\textbf{43.60}\textbf{$\pm$}\scriptsize{2.8 } &\textbf{61.02}\textbf{$\pm$}\scriptsize{1.8}  &\textbf{46.89}  &  
&74.28\textbf{$\pm$}\scriptsize{2.7}  &\textbf{62.78}\textbf{$\pm$}\scriptsize{2.6}  &\textbf{97.35}\textbf{$\pm$}\scriptsize{0.8}  &72.17\textbf{$\pm$}\scriptsize{3.1}  &76.64  \\ \bottomrule
\end{tabular}%
}
\end{center}
\end{table*}

\noindent\textbf{FedSB.} 
We propose FedSB to address the challenges of data heterogeneity in FDG. As shown in Fig. \ref{fig:overview}, FedSB operates through two complementary steps. First, it encourages local clients to learn domain-invariant representations by reducing their overconfidence through label smoothing. Second, it promotes a balanced contribution from different clients by utilizing a simple yet novel budgeting technique. The following sections provide a detailed explanation of these approaches.

Each local client has access to a relatively small dataset within a distinct domain. As a result, the trained local models can produce predictions with high confidence within this domain. However, these models lack generalization and perform poorly when presented with data from other domains.
To alleviate this, we introduce a level of controlled uncertainty to the model to prevent local clients from overfitting to domain-specific representations.
To achieve this, we employ a label smoothing technique and replace hard labels with soft labels, 
by altering ground truth labels as follows:
\begin{equation}
    y'_c =
\begin{cases}
1 - \epsilon + \frac{\epsilon}{M}, & \text{if } c = y \\
\frac{\epsilon}{M}, & \text{if } c \neq y
\end{cases},
\end{equation}
where $\epsilon$ is the smoothing coefficient. This formulation ensures that $\sum_{c=1}^My'_c = 1$. 
Consequently, the cross-entropy loss using the smoothed labels can be derived as:
\begin{equation}
    \mathcal{L} = -\left( (1 - \epsilon + \frac{\epsilon}{M}) \log(p_y) + \sum_{c \neq y} \frac{\epsilon}{M} \log(p_c) \right).
\end{equation}
Here, the collecting terms can be rearranged as:
\begin{equation}
    \mathcal{L} = (1 - \epsilon)(\underbrace{-\log(p_y)}_{\text{NLL Loss}}) + \epsilon (\underbrace{-\sum_{c=1}^{M} \frac{1}{M} \log(p_c)}_{\text{Smooth Loss}}).
\end{equation}
In this formulation, the Negative Log-Likelihood (NLL) loss over the target class encourages correct predictions, whereas the Smooth Loss reduces overconfidence by leveraging the incorrect classes (all classes other than the correct target class). 
We apply $\epsilon$ to control the level of smoothness of the labels, where higher $\epsilon$ values penalize the local models more for overconfident predictions.

Next, let us assume a local training process at round $t+1$ in client $\mathcal{C}_i$. The model is initialized with the global parameters of round $t$ denoted by $\Theta^{t}$. We can characterize the parameters of $\mathcal{C}_i$ after local training, using 
gradient descent, as:  
\begin{equation}
    \theta_i^{t+1} = \Theta^{t} - \eta \sum_{j=1}^{\lfloor \frac{|\mathcal{D}_i|}{B}\rfloor }\nabla \mathcal{L}^j_i,
\end{equation}
where $\eta$ is the learning rate, $B$ is the batch size, and $\mathcal{L}^j_i$ is the loss of the $j^{th}$ batch of the local dataset $\mathcal{D}_i$. By aggregating all the local models in the server using a simple technique such as averaging, the global model is obtained as:
\begin{equation}
    \Theta^{t+1}=\frac{1}{K}\sum_{i=1}^K (\theta_i^{t+1})=\Theta^{t} - \frac{\eta}{K}\sum_{i=1}^K\sum_{j=1}^{\lfloor \frac{|\mathcal{D}_i|}{B}\rfloor }\nabla \mathcal{L}^j_i.
\end{equation}
Accordingly, since
\begin{equation}
\mathbb{E}[\sum_{j=1}^{\lfloor \frac{|\mathcal{D}_i|}{B}\rfloor }\nabla \mathcal{L}^j_i] = \lfloor \frac{|\mathcal{D}_i|}{B}\rfloor \mathbb{E}[\mathcal{L}_i],
\end{equation}
where $\mathbb{E}[\nabla\mathcal{L}_i]$ represent the expected update of client $i$, the expected global model parameters can be described as:
\begin{equation}
\label{eq6}
\mathbb{E}[\Theta^{t+1}] = \Theta^{t} - \frac{\eta}{K}  \sum_{i=1}^K \lfloor \frac{|\mathcal{D}_i|}{B}\rfloor \mathbb{E}[\nabla\mathcal{L}_i].
\end{equation}
According to Eq. \ref{eq6}, we can deduce that clients with larger datasets, i.e., larger $|\mathcal{D}_i|$, have a greater influence in determining the expected value of the global model's update compared to clients with smaller datasets. This can degrade generalization, as the global model tends to gravitate toward the domain of the more influential clients.
To address this issue, we apply a simple 
trick to ensure that each client operates under a fixed training budget regardless of the local dataset size. Specifically, we use a fixed budget for all clients denoted as $\mathcal{S}$. If $|\mathcal{D}_i|> \mathcal{S}$, we randomly select $\mathcal{S}$ samples from $\mathcal{D}_i$, whereas where $|\mathcal{D}_i|< \mathcal{S}$ we oversample $\mathcal{D}_i$. By using the under/over-sampled dataset $\hat{\mathcal{D}}_i$ for training each client, 
the expected global model is derived as:
\begin{equation}
\mathbb{E}[\Theta^{t+1}] = \Theta^{t} - \frac{\eta\mathcal{S}}{KB}  \sum_{i=1}^K \mathbb{E}[\nabla\hat{\mathcal{L}}_i],
\end{equation}
were $\hat{\mathcal{L}}$ denotes the loss over $\hat{\mathcal{D}}_i$. 
As demonstrated in this equation, this straightforward and intuitive solution can effectively ensure equal contribution from all clients toward the global update and mitigate the impact of data heterogeneity, particularly with respect to dataset size.


\section{Experiments}
\textbf{Datasets.}
We evaluate our method on four datasets 
namely PACS 
\cite{li2017deeper}
, OfficeHome 
\cite{venkateswara2017deep}
, VLCS 
\cite{fang2013unbiased}
, and TerraInc 
\cite{beery2018recognition}
. The PACS dataset comprises 9,991 images of the `Photo', `Art Painting', `Cartoon', and `Sketch' domains, each containing seven classes. OfficeHome also has four distinct domains: `Art', `Clipart', `Product', and `Real-world', comprising over 15,500 images belonging to 65 classes. Furthermore, Terra Incognita includes 24,788 pictures of animals from four distinct locations. Finally, VLCS comprises  10,729 images from four distinct object classification datasets: VOC, LabelMe, Caltech101, and SUN09 with five shared classes.


\begin{figure}[t]
    \centering
    {\includegraphics[width=0.44\linewidth]{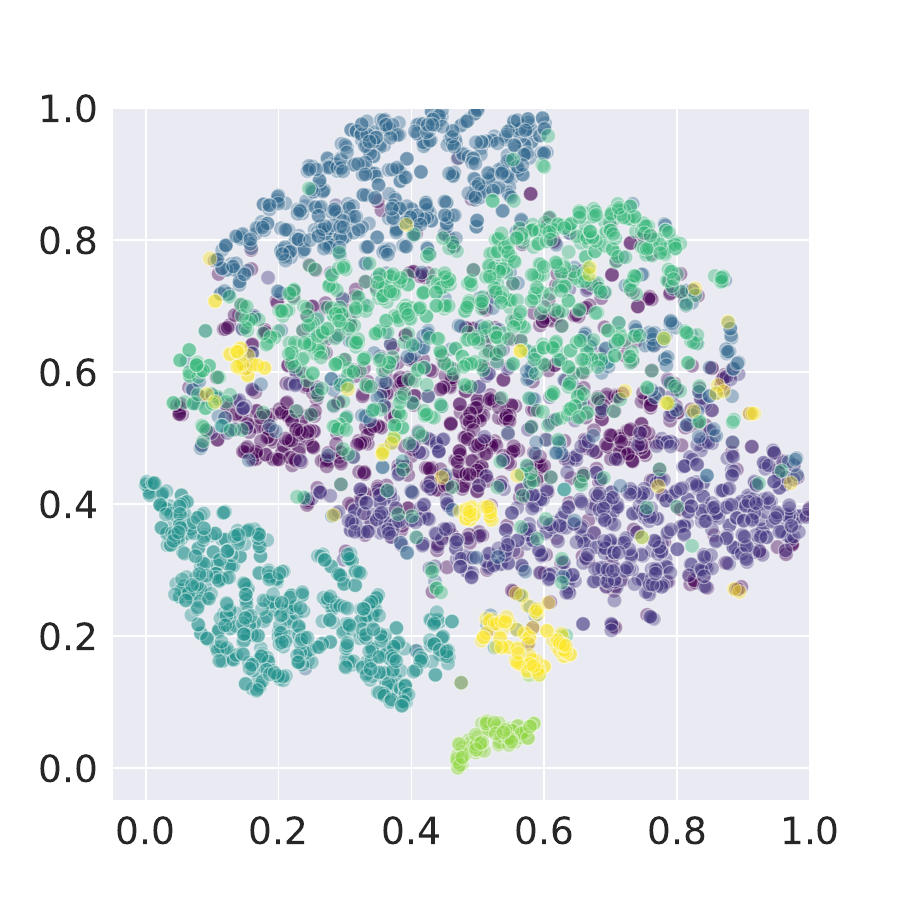}\label{gamma1}}
    \hspace{0.5 cm}
    {\includegraphics[width=0.44\linewidth]{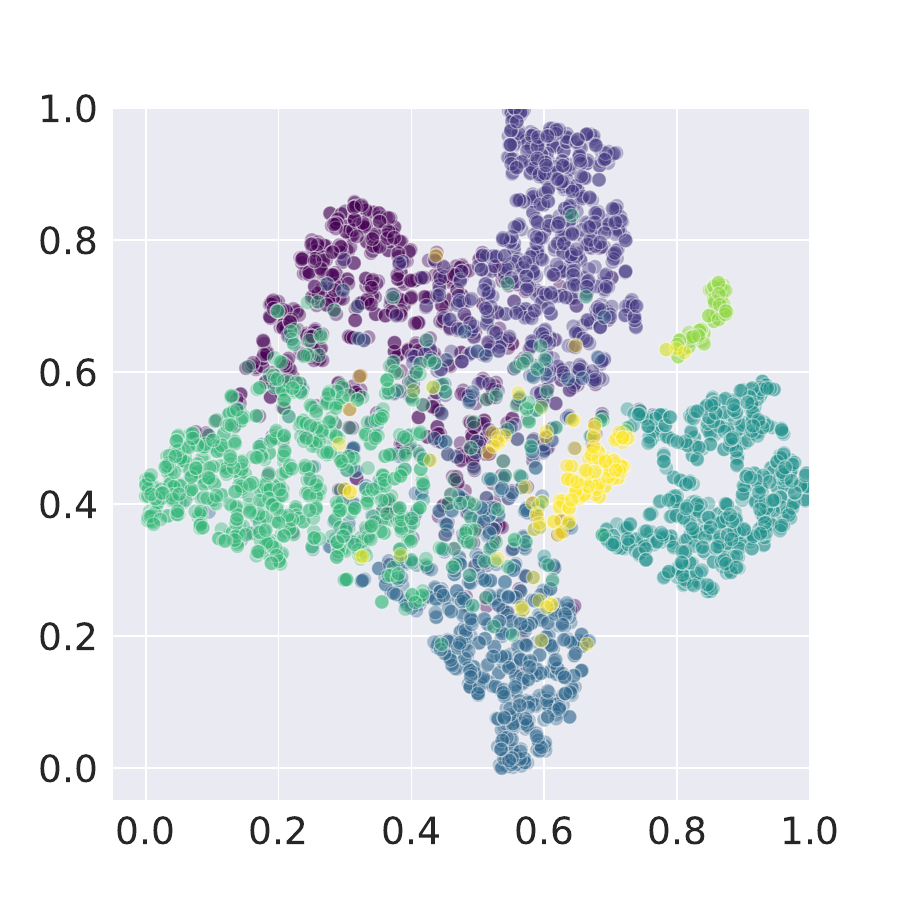}\label{gamma2}}
    \\
    \small{a) FedAVG \hspace{3 cm} b) FedSB}
    \hspace{3 cm}
    \caption{
    TSNE plot for $S$ domain on PACS. The points are color-coded to represent different classes.
    }
    \label{fig:tsne}
\end{figure}

\begin{table}[t]
\caption{Accuracy on PACS using a ResNet-50 backbone.
}
\vspace{5pt}
\begin{center}
\resizebox{0.8\columnwidth}{!}{
\begin{tabular}{lccccc}
\toprule
Method &  P & A & C & S & Ave.\\
\midrule 
FedAvg \cite{mcmahan2017communication} & 95.51  & 82.23   & 78.20  & 73.56 & 82.37 \\ 
FedDG \cite{liu2021feddg} & 96.23  &83.94   & 79.27  & 73.30 & 83.19 \\ 
CCST \cite{chen2023federated}  & \textbf{96.65} & \textbf{88.33} & 78.20 & 82.90 & 86.52 \\ 
FebSB (ours)   & 95.97  & 87.46   & \textbf{80.86} & \textbf{85.22}  & \textbf{87.33}    \\ 
\bottomrule
\end{tabular}
}
\end{center}
\label{Tab:fedsb_vs_ccst}
\end{table}



\begin{table}[t]
\caption{Accuracy of FedAvg and FedSB on the PACS dataset using ViT backbones.}
\vspace{5pt}
\label{tab:fedsb_vit_pacs}
\centering
\resizebox{0.95\columnwidth}{!}{
\begin{tabular}{lcccccc}
\toprule
Backbone & Method & P & A & C & S & Ave.\\
\midrule
\multirow{2}{*}{ViT-b/16} 
& FedAvg \cite{mcmahan2017communication} & \textbf{98.88} & 90.65 & 76.56 & 82.89 & 87.24 \\
& FebSB (ours)   & 98.84 & \textbf{91.49} & \textbf{80.43} & \textbf{83.48} & \textbf{88.56} \\ 
\midrule
\multirow{2}{*}{ViT-b/32} 
& FedAvg \cite{mcmahan2017communication} & 96.66 & 84.57 & 77.46 & \textbf{79.01} & 84.47  \\ 
& FebSB (ours)   & \textbf{96.96} & \textbf{87.89} & \textbf{78.07} & 78.43 & \textbf{85.97} \\ 
\bottomrule 
\end{tabular}
}
\end{table}

\begin{table}[t]
\caption{Accuracy of FedAvg and FedSB on the OfficeHome dataset using ViT backbones.}
\vspace{5pt}
\label{tab:fedsb_vit_officehome}
\centering
\resizebox{0.95\columnwidth}{!}{
\begin{tabular}{lcccccc}
\toprule
Backbone & Method & P & A & C & R & Ave.\\
\midrule
\multirow{2}{*}{ViT-b/16} 
& FedAvg \cite{mcmahan2017communication} & 81.92 & 74.63 & 60.76 & 83.76 & 75.27\\ 
& FebSB (ours)   & \textbf{82.46} & \textbf{75.17} & \textbf{60.80} & \textbf{84.52} & \textbf{75.73} \\ 
\midrule
\multirow{2}{*}{ViT-b/32} 
& FedAvg \cite{mcmahan2017communication} & \textbf{74.93} & 64.82 & 57.47 & 77.33 & 68.64\\ 
& FebSB (ours)   & 74.57 & \textbf{65.65} & \textbf{58.87} & \textbf{78.36} & \textbf{69.36}\\ 
\bottomrule
\end{tabular}
}
\end{table}

\begin{table}[t]
\caption{Accuracy comparison of FedSB and its ablated variants on the PACS dataset}
\vspace{5pt}
\centering
\resizebox{0.9\columnwidth}{!}{
\begin{tabular}{cc|ccccc}
\toprule
Smoothing & Budget & P & A & C  &S & Ave.\\
\midrule 
 -- &--  & 91.67 & 79.25 & 70.46 & 75.98 & 79.34 \\
 -- & \checkmark & 93.93 & 80.25 & 76.63 &77.22 & 81.51 \\
 \checkmark &--  & 93.51 &\textbf{ 82.83} & 75.14 & 82.77 & 83.31 \\
 \checkmark & \checkmark & \textbf{94.19}  & {81.80} & \textbf{75.28} &\textbf{83.52}  & \textbf{83.81} \\ 
\bottomrule
\end{tabular}
}
\label{Tab:ablation}
\end{table}

\begin{table}[t!]
\caption{Impact  of varying $\epsilon$ and $\mathcal{S}$ on the performance of FedSB.}
\vspace{4pt}
\begin{center}
\setlength{\tabcolsep}{8pt}
\resizebox{0.8\columnwidth}{!}{
\begin{tabular}{lccccccc}
\toprule
$variable$ &  P & A & C & S & Ave.\\
\midrule 
$\epsilon=0.1$    &\textbf{93.51}  & \textbf{82.83} & 75.14 & \textbf{82.77} & \textbf{83.31} \\
$\epsilon=0.3$    &93.41  & 81.85 & \textbf{75.73} & 82.39 & 83.35 \\
$\epsilon=0.2$    &93.07  & 80.76 & 75.44 & 82.63 & 82.98 \\
\midrule 
$\mathcal{S}=30B$    &91.34 &80.25 &\textbf{76.63} &77.22  & \textbf{81.51} \\
$\mathcal{S}=45B$    &\textbf{92.40} &\textbf{81.12} &72.08 &75.48 & 80.27 \\
$\mathcal{S}=60B$    &91.87 &80.61 &73.70 &\textbf{78.02} & 81.05 \\
\bottomrule
\end{tabular}
}
\end{center}
\label{Tab:sensitivity}
\end{table}

\noindent \textbf{Evaluation.}
Following \cite{gulrajani2021in}, we utilize the leave-one-domain-out setting. Each time, we select one domain $D_i$ as a target and train the model on the rest of the domains. We repeat this for all domains and average the performance.

\noindent \textbf{Baselines.} We compare FedSB with FedAVG \cite{mcmahan2017communication}, FedADG \cite{zhang2021federated}, FedProx \cite{li2020federated}, FedSR \cite{nguyen2022fedsr}, and FedIIR \cite{guo2023out}. We re-run all the baselines and report the average performance over three different runs, except FedADG \cite{zhang2021federated}, where we report the results from the original paper.

\noindent \textbf{Implementation details.}
Following prior works, we use ResNet-18 for the PACS and VLCS datasets as the feature extractor, while for OfficeHome and TerraIncognita, we utilize ResNet-50. In addition to ResNet-based backbones used by prior works, we also report our results on PACS and OfficeHome using Vision Transformers (ViTs) \cite{dosovitskiy2021an}. Given that ViTs have not been used by the baselines, we re-implement the commonly used FedAVG baseline using two variants of ViTs, namely ViT-b/16 and ViT-b/32. Finally, to compare against CCST \cite{chen2023federated}, we also evaluate our method on PACS with the ResNet-50 backbone. All models are trained with Adam optimizer and a learning rate of 1e-4, using a batch size of 64. We implement and train our method and all of the baselines using the PyTorch framework on an Nvidia RTX 3090 GPU with 24 GB VRAM.

\noindent \textbf{Results.}
Table \ref{tab:results} reports the performance of FedSB in comparison to the baselines on PACS, OfficeHome, TerraINC, and VLCS datasets. As demonstrated, FedSB achieves state-of-the-art results on three out of four datasets. Additionally, as shown in Fig. \ref{fig:tsne}, our method shows better separability of classes compared to that of the FedAVG baseline. Similarly, comparisons against CCST in Table \ref{Tab:fedsb_vs_ccst} demonstrate the superiority of our approach. Lastly, evaluations using ViT backbones are presented in Tables \ref{tab:fedsb_vit_pacs}, and \ref{tab:fedsb_vit_officehome}, where we observe that FedSB continues to outperform the baseline.

Table \ref{Tab:ablation} presents the result of our ablation study where we systematically remove different components of FedSB and report the results on PACS. We observe that the performance drops with the removal of each component, demonstrating the effectiveness of each component in our method. 
We also report the effect of varying $\epsilon$, and 
$\mathcal{S}$
on the overall performance in Table \ref{Tab:sensitivity}. It can be seen that our method does not exhibit high sensitivity to either hyper-parameter.

\section{Conclusion}
\label{sec:conclusion}
We introduced FedSB, a powerful solution to address the challenges of data heterogeneity in FDG. FedSB consists of two components that tackle local model overconfidence and client sample imbalance, respectively. 
We demonstrate the effectiveness of our approach through extensive experiments on four well-known domain generalization benchmark datasets, 
advancing the state-of-the-art by outperforming competing FDG methods on three of the four datasets.
We also provide ablation studies as well as sensitivity analysis, which
provides insight into the different components of our method.

\section*{Acknowledgment}
We would like to gratefully thank Geotab Inc., the City of Kingston, and NSERC for their support and collaboration throughout this work.


\bibliographystyle{ieeetr}
\bibliography{refs}

\end{document}